\documentclass[11pt,a4paper]{article}
\usepackage[margin=1in]{geometry}
\usepackage{amsmath,amssymb}
\usepackage{graphicx}
\usepackage{booktabs,tabularx,array}
\usepackage{enumitem}
\usepackage{float}
\usepackage{subfig}
\usepackage{changepage}
\usepackage{rotating}
\usepackage{xcolor}
\usepackage{tikz}
\usepackage[hidelinks]{hyperref}
\usepackage{microtype}
\newlength{\extralength}
\newlength{\fulllength}
\usetikzlibrary{arrows.meta,positioning,fit,calc}
\title{From Proxy Learning to Driving Decisions: A Transfer-Based Framework for Evaluating Future-Aware Autonomous Driving Planners}
\author{Yikai Wu\\
School of Automation, Nanjing University of Science and Technology\\
Nanjing, China\\
\texttt{wuyikai@njust.edu.cn}}
\date{}
\begin{document}
\maketitle

\begin{abstract}
Future-aware representations and world models are increasingly used in proposal-based autonomous-driving planners to improve trajectory selection. However, improvements in proxy objectives or restricted subsets are often interpreted as planning gains without verifying proposal ordering, selected trajectories, full-scale utility, and critical driving components. We propose the Proxy-to-Decision Transfer (PDT) Framework, an analysis framework that evaluates when learned future information supports a reliable driving-performance improvement claim. Its Decision-Transfer Decomposition Module localizes value loss through score margins, switch-conditioned utility, and support-versus-selection regret. Its Reliability-Constrained Validation Module requires exact pairing, a minimum meaningful effect, scale-expanded confirmation, safety non-compensation, sequential comparability, and family-level robustness. On a representative future-aware planner evaluated with NAVSIM-v1, component BCE decreases from 0.705 to 0.530 while held selected PDM decreases from 0.963 to 0.961. A separate candidate improves a 512-record prefix by 0.00909, with a scene-bootstrap 95\% interval of [0.000744, 0.0177], but its 2048-record and complete-support intervals include zero. A proposal-level replay further confirms the switch-utility decomposition, yet none of 432 screened configurations passes the two-half, two-seed robustness gate. PDT therefore identifies where decision transfer fails or remains indeterminate across proxy, subset, aggregate, and selection evidence.
\end{abstract}

\noindent\textbf{Keywords:} autonomous driving planning; future-aware planning; evaluation framework; proxy-to-decision transfer; decision utility; safety-aware validation; NAVSIM

\section{Introduction}
\label{sec:introduction}

End-to-end autonomous driving has moved beyond isolated perception and motion modules toward systems organized around the final planning objective. Planning-oriented architectures connect perception, prediction, interaction modeling, and trajectory generation within a single computational graph \cite{hu2023uniad,jiang2023vad}. In parallel, driving world models and predictive latent representations have begun to anticipate action-conditioned scene evolution, either to supervise a planner or to evaluate alternative actions \cite{wang2024drivewm,li2025law,zheng2025world4drive,li2025wote}. More recently, joint-embedding predictive learning has been integrated into multimodal proposal-based planning, providing a representative predictive model basis for studying how latent future states affect candidate trajectories \cite{wang2026drivejepa}. These developments make future information increasingly accessible to the planner. The benefit of this information depends on how it affects the selected trajectory and its safety, progress, and comfort.

Evaluation has not advanced at the same granularity as this architectural integration. NAVSIM replaces conventional displacement-only evaluation with scalable non-reactive simulation and a planning-oriented driving score \cite{dauner2024navsim}; other studies have shown that open-loop metrics may reward shortcuts or produce conclusions that do not carry to more decision-relevant evaluation \cite{li2024egostatus,dauner2023misconceptions,jia2024bench2drive}. These benchmarks improve the final measurement target, but they do not by themselves reveal how an internal future-learning gain reaches that target. In a proposal-based planner, a lower future-prediction loss, a better component classifier, a positive development subset, and a higher aggregate driving score are evidence at different levels. Treating them as interchangeable can support a positive claim even when the selected trajectory is unchanged, the gain disappears at full scale, or safety-related components regress. This concern is consistent with the broader predict-then-optimize literature, which distinguishes predictive accuracy from the utility of the induced decision \cite{elmachtoub2022spo,wilder2019decision,mandi2022ranking}.

The challenge is that future-aware planning consists of multiple stages: a learned representation produces proposal scores, the scores select a candidate through a discrete top-1 operation, and the selected candidate determines driving utility. A representation may improve its proxy task without producing utility-aligned proposal scores. Scores may change without crossing the top-1 decision boundary. A well-calibrated selector cannot recover a trajectory that is absent from the proposal set. Finally, a local or aggregate gain may fail under a full, exactly paired, safety-sensitive evaluation. This motivates our study of the observable conditions under which improved future prediction changes the selected trajectory and yields a reliable, non-compensatory improvement in driving utility. The analysis must specify what is measured at each stage, where transfer fails, and when the retained data are insufficient for a verdict.

We address this problem with the Proxy-to-Decision Transfer (PDT) Framework. PDT organizes evidence into five testable failure locations---representation, scoring, selection, support, and deployment---and contains two complementary modules. The Decision-Transfer Decomposition Module uses score margins, an exact switch-conditioned utility identity, and a support-versus-selection regret identity to localize the loss of decision value. The Reliability-Constrained Validation Module then determines whether the observed effect exceeds a declared meaningful-effect threshold and survives exact pairing, scale expansion, safety non-compensation, sequential metric comparability, and fixed statistical units. We evaluate the framework on a representative joint-embedding future-aware planner with NAVSIM-v1 \cite{wang2026drivejepa,dauner2024navsim}. The evidence includes a component proxy that improves from 0.705 to 0.530 while held selected-trajectory utility declines, positive locked-prefix effects that reverse on a complete 12,146-record evaluation pass, aggregate scores that are indistinguishable at three significant digits despite regressions in several critical components, and a 216-arm calibration screen in which no checkpoint improves both fixed validation halves across both seeds. These tests examine whether PDT separates a learned proxy from a reliable driving decision. Their empirical scope is limited to this planner and benchmark.

The contributions of this work are threefold:
\begin{enumerate}[leftmargin=*,label=(\arabic*)]
\item We introduce PDT, a domain-specific analysis and validation framework that turns a future-aware planning claim into an auditable chain of representation, scoring, selection, support, and deployment tests.
\item We operationalize two diagnostic modules: a Decision-Transfer Decomposition that separates decision coverage, conditional switch utility, support regret, and selection regret; and a Reliability-Constrained Validation protocol that combines exact pairing, scale expansion, uncertainty, safety non-compensation, sequential comparability, and family-level robustness.
\item We provide a bounded empirical case study on a future-aware proposal-based planner and NAVSIM-v1 that documents proxy-to-utility, subset-to-full, aggregate-to-component, and best-configuration-to-robust-verdict reversals. Proposal-level replay closes the switch-utility factorization numerically, while paired scene-bootstrap analysis and exact component deltas show which apparently positive effects remain distinguishable from zero or compatible with a declared non-compensation tolerance.
\end{enumerate}

\section{Related Work}
\label{sec:related}

\subsection{Planning-Oriented End-to-End Autonomous Driving}
\label{subsec:rw_e2e}

Early end-to-end driving systems learned a direct mapping from sensory observations to control commands, with conditional imitation learning using high-level navigation commands to resolve multimodal driving behavior \cite{codevilla2018cil}. Subsequent work introduced explicit planning outputs and intermediate structure so that the learned policy could be inspected and trained around the motion-planning objective. The neural motion planner of Zeng et al. jointly reasoned about perception and interpretable cost-based planning \cite{zeng2019nmp}; MP3 integrated mapping, perception, prediction, and planning in a common representation \cite{casas2021mp3}; and ST-P3 used spatial--temporal features and a learned cost volume to connect scene understanding with trajectory planning \cite{hu2022stp3}. Object-centric and sensor-fusion approaches such as PlanT and TransFuser further showed that structured tokens and multimodal temporal fusion can support closed-loop driving policies \cite{renz2022plant,chitta2023transfuser}. These developments made planning the organizing objective of the end-to-end stack, with perception outputs optimized jointly for that objective \cite{chen2024e2e}.

Recent planning-oriented systems have increasingly adopted integrated or proposal-based formulations. UniAD optimizes perception, prediction, and planning tasks in one network, whereas VAD represents agents and map elements as vectors to reduce dense scene-processing overhead \cite{hu2023uniad,jiang2023vad}. SparseDrive extends this direction with sparse scene representations and hierarchical planning selection \cite{sun2024sparsedrive}. Generative planners then enlarge the action support: DiffusionDrive uses a truncated diffusion process initialized from multimodal anchors to generate diverse trajectories efficiently \cite{liao2025diffusiondrive}. Temporal consistency has also become a specific design target. MomAD introduces planning-query momentum to stabilize successive decisions \cite{song2025momad}, while MFPAD combines an autoregressive LSTM memory branch, a Transformer-based forgetting branch, and gated trajectory correction for long-horizon planning \cite{wu2026mfpad}. MFPAD illustrates an architecture-level method for improving temporal planning. Our study examines how an internal predictive signal should be evaluated before it is interpreted as a driving-decision improvement.

The dominant unit of comparison in this literature remains the complete planner: a new representation, decoder, memory mechanism, or proposal generator is judged by its final benchmark score. This is appropriate for system ranking, but it leaves the transfer mechanism underdetermined. When a planner contains a future auxiliary loss, a learned scorer, and a discrete top-1 selector, an end-score difference alone cannot reveal whether the gain arose from a better representation, a changed proposal ordering, a richer candidate set, or an evaluation artifact. PDT complements architecture-centered research by testing these locations separately and avoiding default attribution of the final score to the named module.

\subsection{Future-Aware Representation and World-Model-Based Planning}
\label{subsec:rw_world}

Future information enters autonomous-driving systems in several distinct roles. One line predicts task-specific spatial states that are directly useful to a planner. Self-supervised freespace forecasting, for example, predicts future navigable space to support safe local motion planning \cite{hu2021freespace}. DriveWorld pre-trains spatiotemporal scene representations by modeling dynamic and static evolution, and OccWorld predicts future 3D occupancy to learn scene dynamics \cite{min2024driveworld,zheng2024occworld}. In these methods, future prediction primarily acts as representation supervision: the learned state is expected to improve downstream tasks, but it need not be queried as an explicit simulator when a trajectory is selected.

A second line develops generative driving world models. DriveDreamer learns controllable real-world driving-video generation, GAIA-1 models video, text, and actions autoregressively, and Vista emphasizes high-fidelity generation with versatile control \cite{wang2023drivedreamer,hu2023gaia,gao2024vista}. Drive-WM further demonstrates action-conditioned multiview forecasting and uses predicted futures to compare driving alternatives \cite{wang2024drivewm}. These works expand the ability to imagine scene evolution, yet visual fidelity or future-prediction accuracy is not itself a decision criterion: the generated future must still be converted into a score that orders feasible trajectories according to driving utility.

A third line couples latent future prediction more tightly to planning. LAW uses a latent world model as an auxiliary signal for end-to-end driving \cite{li2025law}. World4Drive predicts intention-conditioned latent futures and uses a world-model selector to evaluate multimodal trajectories \cite{zheng2025world4drive}; WoTE learns online trajectory evaluation from a BEV world model \cite{li2025wote}; and joint-embedding predictive learning has been combined with multimodal trajectory distillation \cite{wang2026drivejepa}. More recently, ProDrive has coupled a query-centric planner with an action-conditioned BEV world model so that candidate trajectories and predicted scene evolution can be assessed jointly \cite{fu2026prodrive}. These systems bring the future representation closer to the action-selection boundary. Nevertheless, the usual evaluation still compares proxy losses, final selected trajectories, or aggregate scores separately. The unresolved methodological issue is whether an observed proxy improvement changes relative proposal scores enough to cross that boundary, whether the resulting switch has positive conditional utility, and whether the candidate set contains a better action at all. PDT formalizes these questions through margin, switch-utility, and support-regret analyses.

\subsection{Planning Evaluation and Decision-Oriented Validation}
\label{subsec:rw_eval}

Planning evaluation now extends beyond open-loop trajectory matching to interactive and decision-relevant protocols. The nuScenes dataset became a common basis for open-loop planning comparisons \cite{caesar2020nuscenes}, whereas nuPlan provides a large-scale closed-loop planning benchmark \cite{caesar2021nuplan} and Bench2Drive evaluates multiple driving abilities in closed-loop CARLA scenarios \cite{jia2024bench2drive}. NAVSIM occupies a complementary position: it uses real logs and a non-reactive simulator to compute a planning-oriented driving score at scale \cite{dauner2024navsim}. Generative and pseudo-simulation platforms further seek greater feedback realism or broader log reuse \cite{yang2024drivearena,cao2025pseudosim}. Together, these benchmarks provide different trade-offs among realism, interactivity, scalability, and reproducibility; consequently, a result is interpretable only relative to the estimand and state assumptions of the chosen evaluator.

Several studies have exposed failure modes in common planning comparisons. Analyses on nuScenes show that ego-state shortcuts and metric definitions can dominate apparent open-loop performance \cite{li2024egostatus,zhai2023rethinking}. Dauner et al. further distinguish vehicle-motion-planning quality from commonly conflated prediction-style metrics and evaluation conventions \cite{dauner2023misconceptions}. These findings motivate stronger final metrics, but PDT addresses an additional level of validity. Even with a decision-oriented benchmark, comparisons can be distorted by unmatched supports, prefix-based selection, history-dependent evaluator state, compensating safety regressions, or repeated checkpoint search. The framework therefore separates the quality of the benchmark from the reliability of the evidence produced with it.

The broader predict-then-optimize and decision-focused learning literature motivates this separation. Smart Predict-then-Optimize directly targets decision regret \cite{elmachtoub2022spo}; differentiable optimization layers and ranking-based formulations train predictions according to downstream decisions rather than predictive error alone \cite{wilder2019decision,mandi2022ranking}. Adaptive data analysis also shows why repeatedly consulting a holdout set can invalidate an apparently favorable selection \cite{dwork2015adaptive}. PDT applies these principles to proposal-based autonomous-driving planners. It links four concerns in one evaluation method: proxy-to-score transfer, score-to-selection activation, support-versus-selection regret, and reliability gates for paired, safety-sensitive, sequential evaluation.

Table~\ref{tab:positioning} makes this distinction explicit. Architecture papers primarily ask whether a complete planner improves; decision-focused learning asks how a predictive model can be optimized for a downstream decision; and benchmark-validity studies ask whether the final evaluator measures a relevant planning outcome. PDT instead audits whether evidence is licensed to move between these levels. It can therefore be used with an unchanged planner and evaluator.

\begin{sidewaystable}[p]
\caption{Positioning of PDT relative to adjacent research strands. A check mark denotes a primary, operationalized concern; topics mentioned only as limitations are left unchecked.\label{tab:positioning}}
\small
\begin{tabularx}{\textheight}{p{4.0cm}p{3.0cm}p{3.0cm}p{3.5cm}p{3.2cm}X}
\toprule
\textbf{Research strand} & \textbf{Proxy--decision alignment} & \textbf{Argmax activation} & \textbf{Support/selection separation} & \textbf{Safety non-compensation} & \textbf{Scale, sequence, and multiplicity gates}\\
\midrule
Planning-architecture comparisons & Indirect & Rarely explicit & Architecture-dependent & Final components may be reported & Benchmark protocol dependent\\
Decision-focused learning & \(\checkmark\) & Optimization dependent & Usually fixed feasible set & Not domain-specific & Not its primary scope\\
Planning-benchmark validity & Final-outcome focus & Not internalized & Sometimes via oracle analysis & Metric dependent & \(\checkmark\)\\
PDT framework & \(\checkmark\) & \(\checkmark\) & \(\checkmark\) & \(\checkmark\) & \(\checkmark\)\\
\bottomrule
\end{tabularx}
\end{sidewaystable}

\section{Proposed Method}
\label{sec:method}

\subsection{Overall PDT Framework}
\label{subsec:overall}

PDT evaluates an improvement claim for a proposal-based planner; it does not prescribe a new planner architecture. Let \(x_i\) denote scene \(i\), and let \(\mathcal{C}_i=\{\tau_{i1},\ldots,\tau_{iK_i}\}\) be the candidate trajectories available to the frozen planning interface. A baseline system produces a future-related representation \(z_i^{(0)}\), proposal scores \(s_{ik}^{(0)}\), and a selected index
\begin{equation}
a_i^{(0)}=\arg\max_{k\in\{1,\ldots,K_i\}}s_{ik}^{(0)}
\label{eq:baseline-selection}
\end{equation}
A candidate intervention produces \(z_i^{(1)}\), \(s_{ik}^{(1)}\), and \(a_i^{(1)}\). The benchmark utility of trajectory \(\tau_{ik}\) is denoted by \(u_i(\tau_{ik})\). It may be an aggregate planning score, but the corresponding component vector \(\mathbf{q}_i(\tau_{ik})\) is retained whenever the claim concerns safety or driving quality. All paired effects are defined only on a common, finite, evaluator-equivalent support \(\mathcal{I}\).

Within this interface, PDT identifies five failure locations:
\begin{enumerate}[leftmargin=*,label=\textbf{F\arabic*.}]
\item \textbf{Representation transfer failure}: the held future/proxy objective does not improve, or its improvement is not stable under the declared split.
\item \textbf{Scoring transfer failure}: the representation improves, but representation-derived scores do not better order candidates by the declared driving utility.
\item \textbf{Selection transfer failure}: scores change without changing the top-1 trajectory, or the changed trajectories have non-positive conditional utility.
\item \textbf{Support transfer failure}: higher-utility trajectories are not present in \(\mathcal{C}_i\), so score calibration cannot remove the remaining regret.
\item \textbf{Deployment transfer failure}: a local effect does not survive the full support, critical metric vector, sequential evaluator state, or a locked robustness criterion.
\end{enumerate}
A stage is marked \textsc{pass}, \textsc{fail}, or \textsc{indeterminate}. Missing proposal arrays, non-comparable evaluator states, or unmatched scenes produce \textsc{indeterminate}, not a negative result.

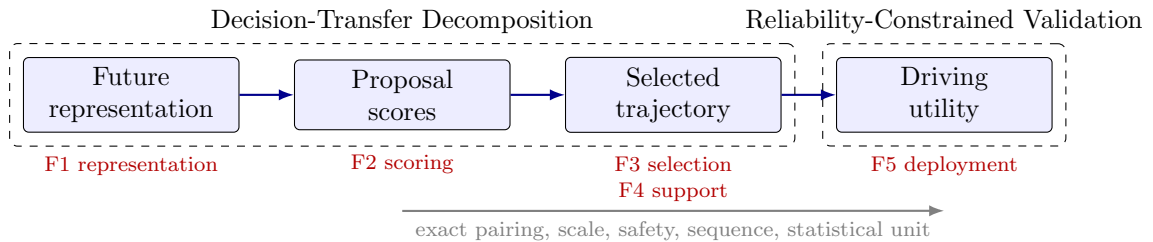
\begin{figure}[H]
\centering
\begin{tikzpicture}[
  node distance=0.72cm,
  stage/.style={draw, rounded corners=2pt, minimum width=2.85cm, minimum height=0.82cm, align=center, font=\small, fill=blue!7},
  fail/.style={font=\scriptsize, text=red!70!black, align=center},
  module/.style={draw, dashed, rounded corners=3pt, inner sep=5pt},
  arrow/.style={-{Latex[length=2mm]}, thick, blue!55!black}
]
\node[stage] (rep) {Future\\representation};
\node[stage, right=of rep] (score) {Proposal\\scores};
\node[stage, right=of score] (sel) {Selected\\trajectory};
\node[stage, right=of sel] (util) {Driving\\utility};
\draw[arrow] (rep) -- (score);
\draw[arrow] (score) -- (sel);
\draw[arrow] (sel) -- (util);
\node[fail, below=0.18cm of rep] {F1 representation};
\node[fail, below=0.18cm of score] {F2 scoring};
\node[fail, below=0.18cm of sel] {F3 selection\\F4 support};
\node[fail, below=0.18cm of util] {F5 deployment};
\node[module, fit=(rep)(score)(sel), label={[font=\small]above:Decision-Transfer Decomposition}] (m1) {};
\node[module, fit=(util), label={[font=\small]above:Reliability-Constrained Validation}] (m2) {};
\draw[-{Latex[length=2mm]}, thick, gray] ($(m1.south)+(0,-0.85)$) -- node[below,font=\scriptsize]{exact pairing, scale, safety, sequence, statistical unit} ($(m2.south)+(0,-0.85)$);
\end{tikzpicture}
\caption{The Proxy-to-Decision Transfer Framework. The first module diagnoses how learned future information changes scores, decisions, and attainable utility; the second module limits the claim to effects that remain comparable and reliable at deployment scale. The failure labels are diagnostic locations, not mutually exclusive model categories.}
\label{fig:pdt}
\end{figure}

Figure~\ref{fig:pdt} shows how the two modules operate on this chain. The decomposition module asks where decision value is lost. The reliability module asks whether an observed gain is valid at the strength claimed.

\subsection{Decision-Transfer Decomposition Module}
\label{subsec:decomposition}

\subsubsection{Score Margins and Decision-Boundary Activation}

Let \(a_i=a_i^{(0)}\) be the baseline top-1 index and define the score perturbation \(\delta_{ik}=s_{ik}^{(1)}-s_{ik}^{(0)}\). Candidate \(b\neq a_i\) replaces the baseline selection only if
\begin{equation}
\delta_{ib}-\delta_{ia_i}\geq s_{ia_i}^{(0)}-s_{ib}^{(0)}
\label{eq:crossing}
\end{equation}
The baseline top-1/top-2 margin is
\begin{equation}
m_i=s_{ia_i}^{(0)}-\max_{b\neq a_i}s_{ib}^{(0)}
\label{eq:margin}
\end{equation}
Thus, \(\max_{b\neq a_i}(\delta_{ib}-\delta_{ia_i})<m_i\) is a sufficient no-switch condition. A lower proxy loss or a changed score vector is not evidence of a changed decision unless the perturbation is evaluated relative to the decision margin. PDT therefore requires the baseline and candidate proposal arrays, selected indices, top-1/top-2 margins, and exact candidate identities. If these arrays are not retained, the cause of an unchanged argmax cannot be reconstructed from aggregate losses alone.

\paragraph{Proposition 1 (decision-boundary activation).}
For a fixed candidate set, candidate \(b\) replaces the baseline choice \(a_i\) if and only if its relative score perturbation compensates its baseline score deficit, i.e., Equation~\eqref{eq:crossing} holds and the candidate attains the largest perturbed score. Consequently, a proxy or score change has decision relevance only through its proposal-wise activation surplus
\begin{equation}
h_i=\max_{b\neq a_i}\left[(\delta_{ib}-\delta_{ia_i})-(s_{ia_i}^{(0)}-s_{ib}^{(0)})\right]
\label{eq:surplus}
\end{equation}
A switch occurs exactly when \(h_i\geq0\), apart from the declared tie-breaking rule. This formulation is stronger than comparing a global perturbation norm with the top-1/top-2 margin because it preserves which perturbation belongs to which competing proposal.

\subsubsection{Switch-Conditioned Decision Utility}

For a common support of \(N=|\mathcal{I}|\) scenes, define the paired utility change
\begin{equation}
d_i=u_i(\tau_{i,a_i^{(1)}})-u_i(\tau_{i,a_i^{(0)}})
\label{eq:paired-utility-change}
\end{equation}
and the switch indicator \(I_i=\mathbf{1}[a_i^{(1)}\neq a_i^{(0)}]\). Because \(d_i=0\) when the same candidate is selected under a fixed candidate set and evaluator, the mean decision-utility change is
\begin{equation}
\Delta J=\frac{1}{N}\sum_{i\in\mathcal{I}}d_i
=\underbrace{\frac{1}{N}\sum_{i\in\mathcal{I}}I_i}_{\rho}
\underbrace{\frac{\sum_{i\in\mathcal{I}}I_id_i}{\sum_{i\in\mathcal{I}}I_i}}_{\mu}
=\rho\mu
\label{eq:switch}
\end{equation}
Here, \(\rho\) is the trajectory switch rate and \(\mu\) is the conditional utility of a switch. Equation~\eqref{eq:switch} is an exact diagnostic identity under common candidate support, identical utility computation, and exact trajectory identity. It separates two failure modes that a mean delta conceals: the intervention may rarely cross the boundary (\(\rho\) is small), or it may frequently switch to worse trajectories (\(\mu\leq 0\)). When no switch occurs, \(\Delta J=0\) and \(\mu\) is undefined; PDT reports a zero switch count and leaves the conditional value undefined.

\paragraph{Proposition 2 (switch-utility factorization).}
Under fixed candidate identities and a common evaluator, Equation~\eqref{eq:switch} is exact because every non-switch contributes zero paired utility difference. It follows that \(\Delta J>0\) requires both \(\rho>0\) and \(\mu>0\), whereas either \(\rho=0\) or \(\mu\leq0\) is sufficient to reject a positive decision-transfer claim. This is a diagnostic factorization, not an independence assumption between switching and utility.

\subsubsection{Support and Selection Regret}

A score-only intervention is limited by the trajectories it can select. Let \(\Omega_i\) denote an explicitly declared external comparison space, \(\mathcal{C}_i\subseteq\Omega_i\) the available candidates, and
\begin{equation}
u_i^\star=\sup_{\tau\in\Omega_i}u_i(\tau),\qquad
u_i^{\mathcal{C}}=\max_{\tau\in\mathcal{C}_i}u_i(\tau)
\label{eq:oracle-utilities}
\end{equation}
For a selected trajectory \(\hat{\tau}_i\in\mathcal{C}_i\), PDT defines
\begin{align}
R_{\mathrm{total}} &=\frac{1}{N}\sum_i\left(u_i^\star-u_i(\hat{\tau}_i)\right),\\
R_{\mathrm{support}} &=\frac{1}{N}\sum_i\left(u_i^\star-u_i^{\mathcal{C}}\right),\\
R_{\mathrm{selection}} &=\frac{1}{N}\sum_i\left(u_i^{\mathcal{C}}-u_i(\hat{\tau}_i)\right)
\end{align}
Adding and subtracting \(u_i^{\mathcal{C}}\) gives
\begin{equation}
R_{\mathrm{total}}=R_{\mathrm{support}}+R_{\mathrm{selection}}
\label{eq:regret}
\end{equation}
Equation~\eqref{eq:regret} is a bookkeeping identity and makes no general regret claim. A large \(R_{\mathrm{selection}}\) supports further ranking or calibration work; a large \(R_{\mathrm{support}}\) indicates that score-only tuning cannot close the declared gap. If \(\Omega_i\) or proposal-wise utilities are unavailable, PDT reports only the observable term and marks the remaining decomposition indeterminate. Repeated failure of several calibrators is consistent with a support limitation, but does not prove that support is the unique bottleneck.

\paragraph{Proposition 3 (intervention ceiling).}
For any score-only intervention that leaves \(\mathcal{C}_i\) fixed, the best attainable mean utility is \(N^{-1}\sum_i u_i^{\mathcal C}\). Hence the intervention can reduce \(R_{\mathrm{selection}}\) but cannot reduce \(R_{\mathrm{support}}\). Observing a small candidate-oracle gap therefore bounds the value available to further score calibration; attributing the residual total regret to support additionally requires an external comparison space \(\Omega_i\).

Together, Propositions 1--3 define an ordered set of claims. Representation evidence can support a learning claim, proposal-wise activation and positive \(\mu\) can support a decision-level improvement claim, and reliability gates can support a deployment claim. No earlier claim logically entails a later one. PDT therefore reports the first failed or indeterminate transfer stage instead of collapsing the chain into one aggregate score.

\subsection{Reliability-Constrained Validation Module}
\label{subsec:reliability}

The second module determines how far an observed effect may be generalized. A reliable positive decision-transfer verdict requires all applicable conditions below.

\paragraph{Exact pairing.}
Baseline and candidate results must be joined by an immutable scene key. Their evaluated token sets, candidate identities, evaluator version, and finite-value filters must match. The primary estimate is the mean of paired differences, not the difference between independently filtered means.

\paragraph{Scale-expanded confirmation.}
Checkpoint and hyperparameter selection are restricted to a declared development support. A later prefix, unseen suffix, fixed quartiles, or full benchmark may confirm or reject the locked candidate but must not reopen selection. If a full run is authorized after an observed prefix, PDT records it as a diagnostic expansion and does not treat it as an untouched final test.

\paragraph{Safety non-compensation.}
Let \(\Delta \bar{J}\) be the aggregate utility change and let \(\Delta\bar{\mathbf{q}}=(\Delta\bar{q}_1,\ldots,\Delta\bar{q}_L)\) contain preregistered critical components, with tolerances \(\boldsymbol{\epsilon}\geq 0\). Let \(\eta_J>0\) be a preregistered minimum effect that is large enough to be practically meaningful at the benchmark's reporting precision. A reliable aggregate-improvement claim requires
\begin{equation}
\Delta\bar{J}\geq\eta_J
\quad\text{and}\quad
\Delta\bar{q}_\ell\geq-\epsilon_\ell
\quad\forall \ell\in\mathcal{S},
\label{eq:safety}
\end{equation}
where \(\mathcal{S}\) is the declared safety-critical set. If \(\eta_J\) was not fixed before evaluation, PDT reports the observed magnitude as nominal and does not promote a sign-only difference to a reliable improvement. Equation~\eqref{eq:safety} is a claim policy, not a real-world safety certificate. A candidate may show a positive numerical difference while failing both practical-effect and non-compensation requirements.

\paragraph{Sequential comparability.}
History-dependent metrics are compared only when the same current scenes, predecessor records, ordering, state initialization, reset policy, and missing-history rules are used. If these conditions differ, even numerically similar metric names denote different estimands. PDT marks the comparison indeterminate and does not use either implementation as a baseline.

\paragraph{Statistical unit and multiplicity.}
The unit of uncertainty is declared before analysis. Scene-level paired intervals are suitable when scenes are the inferential unit; log- or route-level resampling is used when within-log dependence is material. For large configuration screens, PDT reports the number of arms and checkpoints, fixed halves and seeds, and the family-level gate. The nominal best checkpoint is not promoted when it fails the declared robustness rule, consistent with the risks of adaptive holdout reuse \cite{dwork2015adaptive}.

The resulting stage verdicts are:
\begin{itemize}[leftmargin=*]
\item \textsc{pass}: the required estimate is positive and all applicable comparability, scale, safety, and robustness gates pass;
\item \textsc{fail}: comparable evidence exists, but at least one required transfer or reliability gate fails;
\item \textsc{indeterminate}: the claim cannot be evaluated because a load-bearing artifact, support match, evaluator state, or uncertainty unit is unavailable.
\end{itemize}

\subsection{Framework Execution Protocol}
\label{subsec:protocol}

A third party can apply PDT through the following sequence.
\begin{enumerate}[leftmargin=*,label=\textbf{Step \arabic*.}]
\item \textbf{Freeze the claim and utility.} State the future-aware intervention, baseline, final utility, critical components, and intended claim scope.
\item \textbf{Lock supports and selection rules.} Record data splits, scene/log keys, candidate construction, checkpoint eligibility, seeds, expansion gates, and evaluator hashes before opening the corresponding outcomes.
\item \textbf{Test representation transfer.} Evaluate the future/proxy objective on a held support not used for final model choice.
\item \textbf{Test scoring and margins.} Persist proposal identities, baseline/candidate scores, utility labels when available, top-1/top-2 margins, and relative perturbations.
\item \textbf{Test selection and regret.} Report switch count, \(\rho\), \(\mu\), \(\Delta J\), candidate oracle, and observable regret terms.
\item \textbf{Apply reliability gates.} Verify exact pairing, fixed subsets, unseen expansion/full support, the critical component vector, sequential context, and statistical unit.
\item \textbf{Issue a bounded verdict.} Report the first failed transfer stage, every failed reliability gate, and all indeterminate quantities.
\end{enumerate}

The minimum reproducibility record contains scene and log keys; candidate trajectories or stable identifiers; baseline and candidate proposal scores; selected indices; per-candidate utility labels when available; aggregate and component metrics; checkpoint, code, evaluator, and data hashes; random seeds; finite/missing flags; and the exact aggregation script.

\section{Experimental Results}
\label{sec:experiments}

\subsection{Datasets and Evidence Units}
\label{subsec:data}

We instantiate PDT on a representative proposal-based future-aware planner evaluated with NAVSIM-v1. The model basis combines a joint-embedding predictive visual representation with multimodal trajectory proposals \cite{wang2026drivejepa}, while NAVSIM uses real driving logs and non-reactive simulation to compute planning-oriented metrics \cite{dauner2024navsim}. The study is an in-depth case analysis of one model family and one benchmark, not a cross-architecture validation of PDT.

The evidence was produced by a sequence of locked development, prefix, and full-evaluation protocols. Table~\ref{tab:evidence-units} lists the units used in the analysis. The framework was synthesized retrospectively from these experiments, whereas individual expansion thresholds, fixed halves, checkpoint locks, and safety gates were recorded within their campaigns before the relevant outcomes were opened. We do not claim that the complete PDT taxonomy was preregistered.

The full NAVSIM-v1 comparisons generated 12,146 valid output records with zero failed predictions and exact baseline/candidate token support. Because the public benchmark distinguishes required prediction records from the subset or weighting used by a particular scoring release, we describe 12,146 as the complete record support. The reported PDMS values are taken from the locked official evaluator; we do not reinterpret the denominator without the exported evaluator table.

\begin{table}[H]
\caption{Evidence units used to instantiate PDT. Descriptive names indicate the methodological role of each model or experiment; internal run identifiers are omitted.\label{tab:evidence-units}}
\begin{adjustwidth}{-\extralength}{0cm}
\begin{tabularx}{\fulllength}{p{3.4cm}p{3.6cm}Xp{3.6cm}}
\toprule
\textbf{Evidence unit} & \textbf{Support} & \textbf{Purpose in PDT} & \textbf{Primary boundary}\\
\midrule
Decision-boundary audit & 16 training and 4 validation tokens; 32 proposals per token & Candidate-oracle headroom and observed decision-boundary activity & Proposal residual arrays and trained checkpoint were not retained\\
Component-supervised scorer & Held proposal/utility evaluation & Proxy-to-selected-utility alignment & One instantiated scorer; not a universal proxy result\\
Direct-utility candidate & Held support and locked prefix of 512 records & Development-to-prefix transfer under exact scoring batches & Only eight prefix decisions differed\\
Future-supervised candidate & 512/128 teacher windows; prefixes 512 and 2048; complete 12,146-record pass & Learnability of future signal and prefix-to-full deployment transfer & Full run was authorized after prefix inspection and is diagnostic\\
Aggregate-score candidate & Exact paired complete baseline/candidate pass & Aggregate and critical-component verdict & Benchmark evidence is not a real-world safety certificate\\
Calibration stress-test family & 256/64 frozen tokens; 32 proposals; 216 arms; 432 two-seed configurations & Margin, switch-utility, regret, and family-level robustness & External support regret remains unavailable\\
\bottomrule
\end{tabularx}
\end{adjustwidth}
\end{table}

\subsection{Evaluation Metrics}
\label{subsec:metrics}

The experiments use four levels of evidence. Future temporal loss and component binary cross-entropy (BCE) measure whether an internal target is learned. Lower values are better, but they are not treated as driving utility. Where proposal arrays are present, PDT uses ranking/margin diagnostics, selected indices, switch count, \(\rho\), and \(\mu\). When only selected indices remain, the analysis can determine whether a switch occurred but cannot reconstruct the unobserved ranking or perturbation scale.

At proposal level, selected PDM is the utility of the chosen candidate. On the official NAVSIM-v1 evaluation, the PDM score combines no-at-fault collision and drivable-area compliance as multiplicative terms with weighted ego progress, TTC, and comfort:
\begin{equation}
\mathrm{PDM}_i=q_{i,\mathrm{collision}}q_{i,\mathrm{drivable}}
\frac{5q_{i,\mathrm{progress}}+5q_{i,\mathrm{TTC}}+2q_{i,\mathrm{comfort}}}{12}
\label{eq:pdm}
\end{equation}
PDMS is the official aggregation of scenario PDM scores. We report the aggregate together with collision, drivable-area compliance, TTC, driving direction, and comfort directions whenever these fields are available.

The original protocols use exact token pairing, fixed halves or quartiles, unseen suffixes, zero-failure checks, minimum expansion effects, and nonnegative critical-component changes. A gate defines model selection and is distinct from a null-hypothesis significance test. For comparisons whose paired scenario CSVs were retained, we additionally report percentile intervals from 10,000 deterministic paired bootstrap resamples (seed 20260902). Each resample uses the paired scenario difference; the two systems are never resampled independently. The exported CSVs do not contain log identifiers, so these intervals treat scenarios as the resampling unit and do not account for within-log dependence. They summarize uncertainty on the recorded support and do not replace route- or log-level inference. Main-text metrics use approximately three significant digits; smaller deltas use scientific notation, while machine-precision values remain in the reproducibility artifacts.

\subsection{Implementation Details}
\label{subsec:implementation}

The public joint-embedding predictive backbone and the NAVSIM-v1 evaluator were frozen. The full paired protocol byte-compared the scoring implementation and configuration between candidate and clean baseline repositories. Checkpoint selection used outcome-independent development support before official evaluation. Candidate and baseline CSVs were joined by token, required identical token sets, and were checked for finite values and zero failures.

Direct utility targets were regenerated in the exact scoring batches used by the held evaluator; mixed target-cache batching was rejected when it produced small score differences. Fixed validation halves were defined independently of outcomes, and the calibration campaign evaluated both halves across two seeds. Prefix expansion used locked checkpoints and immutable comparison CSVs. History-dependent comfort experiments required an evaluator in which the full result equals the weighted mean of the same two fixed folds; comparisons using different batch/seed histories were rejected as non-decomposable.

The calibration stress test reconstructed an official baseline selected utility of 0.978 from frozen tensors. It trained 216 arms spanning three calibrator forms, three perturbation caps, three regularization strengths, two weight decays, two learning rates, and two seeds. Four evaluation steps per arm produced 864 seed-level outcomes, which were grouped into 432 two-seed configurations for the family-level gate. A CPU replay used the same frozen 256 training tokens, 64 validation tokens, and 32 proposals per token to recover proposal-wise margins, selections, and utilities for representative nominal configurations. All arms were retained in the family-level conclusion.

\subsection{Main Experiments}
\label{subsec:main-results}

\subsubsection{Test 1: Proxy Improvement and Selected Utility}

The component-fitting experiment in Table~\ref{tab:proxy-boundary} provides a direct counterexample. Component BCE decreased from 0.705 to 0.530, while held selected PDM decreased from 0.963 to 0.961. The internal prediction task improved, but the trajectory chosen under the resulting scores had lower mean utility. This rejects the sufficiency claim that better component BCE implies better selected PDM for the tested scorer. It does not show that component supervision is generally harmful.

A separate future-supervision campaign supplies a complementary positive premise. A frozen real-future teacher set of 512 training and 128 validation windows was constructed, and the selected continual-supervision checkpoint improved held common planning loss by 0.0543. Its later positive prefixes show that learning occurred on the retained support. The remaining analysis tests where that signal transfers and over which evaluation support.

\subsubsection{Test 2: Decision-Boundary Activation}

In the small-scale decision-boundary audit, future temporal loss decreased from 0.0527 to 0.0329. Nevertheless, the four validation selections remained \([17,28,21,20]\) at steps 0, 25, and 50, and selected PDM remained 0.978. Candidate labels show 1.42\% oracle headroom on 16 training tokens and 2.20\% on four validation tokens. Nonzero headroom therefore coexisted with zero observed validation switching.

The run did not retain per-proposal adjusted scores, residual arrays, or a trained checkpoint. Consequently, Equation~\eqref{eq:crossing}, rank correlation, and a frozen scale sweep cannot be reconstructed. PDT assigns an observed selection-transfer failure and an \textsc{indeterminate} scoring-versus-magnitude cause; it does not infer that the learned ranking was poor.

A later frozen-tensor calibration screen supplies the proposal-level arrays missing from that early audit. Table~\ref{tab:proposal-diagnostics} reports two representative two-seed configurations at their final evaluation step. The repeatable linear configuration changes four of 64 validation decisions in both seeds, giving \(\rho=0.0625\), \(\mu=0.00170\), and \(\Delta J=1.06\times10^{-4}\). Direct computation and \(\rho\mu\) agree to floating-point precision, and all four switches satisfy the proposal-wise activation condition in Equation~\eqref{eq:surplus}. However, three switches exchange proposals with equal recorded utility and only one produces a positive difference; the interval has a zero lower endpoint, and the second fixed half has zero gain. The result confirms the decomposition but does not show a reliable improvement.

The nominally best set-wise configuration further illustrates seed instability. Seed 0 changes seven decisions and yields \(\Delta J=2.25\times10^{-4}\), whereas seed 1 changes none. Its positive sign therefore cannot be interpreted as replicated decision transfer. In all configurations the candidate-oracle selection regret is observable, but support regret remains \textsc{indeterminate} because the frozen artifacts contain no external trajectory space beyond the 32 proposals.

\begin{table}[H]
\caption{Proposal-level PDT diagnostics from CPU replay of frozen validation tensors. Intervals are 95\% percentile intervals from 10,000 validation-token bootstrap resamples.\label{tab:proposal-diagnostics}}
\begin{adjustwidth}{-\extralength}{0cm}
\begin{tabularx}{\fulllength}{p{3.2cm}p{1.4cm}p{1.8cm}p{1.8cm}p{2.4cm}p{2.7cm}X}
\toprule
\textbf{Configuration} & \textbf{Seed} & \textbf{Switches / 64} & \(\boldsymbol{\rho}\) & \(\boldsymbol{\mu}\) & \(\boldsymbol{\Delta J}\) \textbf{[95\% interval]} & \textbf{PDT verdict}\\
\midrule
Set-wise nominal best & 0 & 7 & 0.109 & 0.00205 & \(2.25\times10^{-4}\) [0, \(6.74\times10^{-4}\)] & Nominal only; 1 positive and 6 utility-neutral switches\\
Set-wise paired run & 1 & 0 & 0 & Undefined & 0 [0, 0] & No selection transfer\\
Linear repeatable & 0 & 4 & 0.0625 & 0.00170 & \(1.06\times10^{-4}\) [0, \(3.19\times10^{-4}\)] & Exact factorization; robustness gate fails\\
Linear repeatable & 1 & 4 & 0.0625 & 0.00170 & \(1.06\times10^{-4}\) [0, \(3.19\times10^{-4}\)] & Same selections; robustness gate fails\\
\bottomrule
\end{tabularx}
\end{adjustwidth}
\end{table}

\begin{figure}[H]
\begin{adjustwidth}{-\extralength}{0cm}
\centering
\subfloat[\centering\label{fig:margin-activation}]{\includegraphics[width=0.55\fulllength]{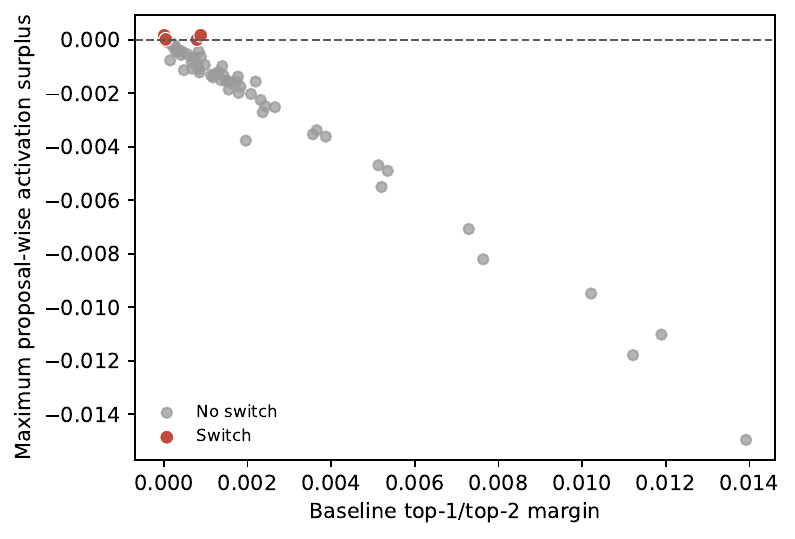}}
\hfill
\subfloat[\centering\label{fig:switch-utility}]{\includegraphics[width=0.40\fulllength]{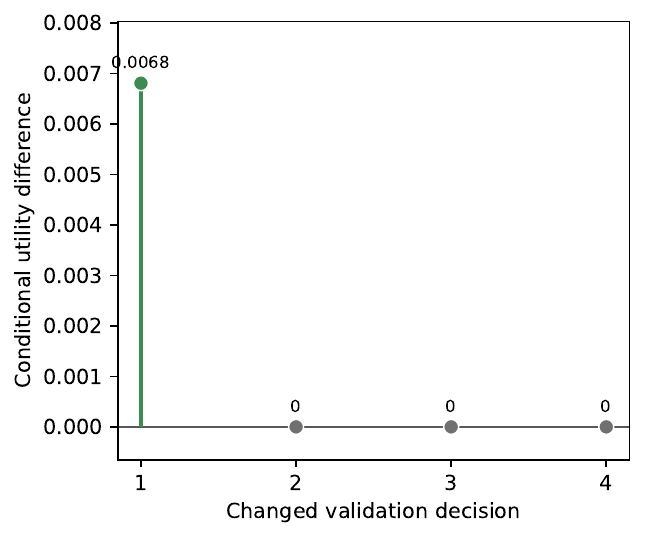}}
\end{adjustwidth}
\caption{Proposal-level decision-boundary diagnostics for one repeatable linear calibration run. (\textbf{a}) Maximum proposal-wise activation surplus against the baseline top-1/top-2 margin; red markers identify the four switched cases. (\textbf{b}) Utility differences for those four changed decisions; the markers and value labels make the three exact zeros visible, while the remaining decision improves by 0.00681.}
\label{fig:margin-switch}
\end{figure}

Figure~\ref{fig:margin-switch} shows the exact activation surplus for the repeatable linear configuration. Only four points cross the switching boundary; their small baseline margins make them susceptible to the learned score perturbation. Figure~\ref{fig:switch-utility} explains why switch count alone is insufficient: the four switched utility differences are 0.00681, 0, 0, and 0, so one changed decision carries the entire positive mean.

\begin{table}[H]
\caption{Proxy and decision-boundary evidence.\label{tab:proxy-boundary}}
\begin{adjustwidth}{-\extralength}{0cm}
\begin{tabularx}{\fulllength}{p{3.1cm}p{3.7cm}p{4.3cm}X}
\toprule
\textbf{Test} & \textbf{Proxy evidence} & \textbf{Decision evidence} & \textbf{PDT verdict}\\
\midrule
Component-supervised scorer & BCE \(0.705\rightarrow0.530\) & Held PDM \(0.963\rightarrow0.961\) & Scoring/selection utility does not inherit the proxy improvement\\
Decision-boundary audit & Future loss \(0.0527\rightarrow0.0329\) & Same four selections and PDM \(0.978\); four-token validation oracle headroom \(2.20\%\) & No observed selection transfer; cause within scoring/magnitude is indeterminate\\
\bottomrule
\end{tabularx}
\end{adjustwidth}
\end{table}

\subsubsection{Test 3: Deployment-Scale Stability of Local Gains}

Exact scoring-batch reconstruction first produced a held selected-PDM difference of \(+6.38\times10^{-4}\), with positive fixed halves and nonnegative available safety deltas. When the locked checkpoint was evaluated on an official 512-record prefix, the difference from the frozen development reference was only \(+1.15\times10^{-4}\), below the preregistered \(5.00\times10^{-4}\) expansion threshold. Seven records improved, 504 tied, and one declined. The paired scene-bootstrap interval is [\(2.19\times10^{-5}\), \(2.45\times10^{-4}\)], but the prespecified practical expansion gate still fails. The result is therefore a precisely estimated small local effect, not evidence for deployment-scale improvement.

A later future-supervised candidate produced \(+0.00909\) over the frozen development reference on prefix512, with a paired interval of [0.000744, 0.0177]. The mean remained positive by \(+0.00250\) on prefix2048, but its interval widened across zero to [\(-3.11\times10^{-4}\), 0.00527]. The unseen suffix of 1536 records remained positive and three of four fixed quartiles were positive, but the strict-quartile and comfort gates failed. The frozen candidate was expanded to a complete 12,146-record pass. All three complete-support PDMS values round to 0.937; at higher precision, the candidate is lower than the frozen development reference by \(6.64\times10^{-4}\), with interval [\(-0.00248\), 0.00121], and lower than the official checkpoint by \(2.20\times10^{-4}\), with interval [\(-0.00236\), 0.00191]. The evidence therefore changes from a locally distinguishable positive mean to an unresolved and slightly negative complete-support mean.

Because prefix2048 authorized the complete run, the full result is a locked diagnostic expansion and cannot serve as an untouched checkpoint-selection test. It shows deployment-transfer failure for the fixed candidate; unbiased evaluation of a newly selected model remains future work.

\begin{table}[H]
\caption{Validation-to-deployment transfer. All deltas name the comparison baseline.\label{tab:scale}}
\begin{adjustwidth}{-\extralength}{0cm}
\begin{tabularx}{\fulllength}{p{3.0cm}p{3.0cm}p{3.1cm}p{3.3cm}X}
\toprule
\textbf{Candidate} & \textbf{Held/local} & \textbf{Prefix512} & \textbf{Prefix2048} & \textbf{Complete evaluation verdict}\\
\midrule
Direct-utility candidate & \(+6.38\times10^{-4}\) held PDM & \(+1.15\times10^{-4}\) [\(2.19\times10^{-5}\), \(2.45\times10^{-4}\)]; gate fail & Not authorized & No expansion; effect coverage 8/512\\
Future-supervised candidate & Held common loss \(+0.0543\) & \(+0.00909\) [0.000744, 0.0177] & \(+0.00250\) [\(-3.11\times10^{-4}\), 0.00527]; comfort/quartile fail & \(-6.64\times10^{-4}\) [\(-0.00248\), 0.00121] vs frozen; \(-2.20\times10^{-4}\) [\(-0.00236\), 0.00191] vs official\\
\bottomrule
\end{tabularx}
\end{adjustwidth}
\end{table}

\begin{figure}[H]
\centering
\includegraphics[width=0.90\textwidth]{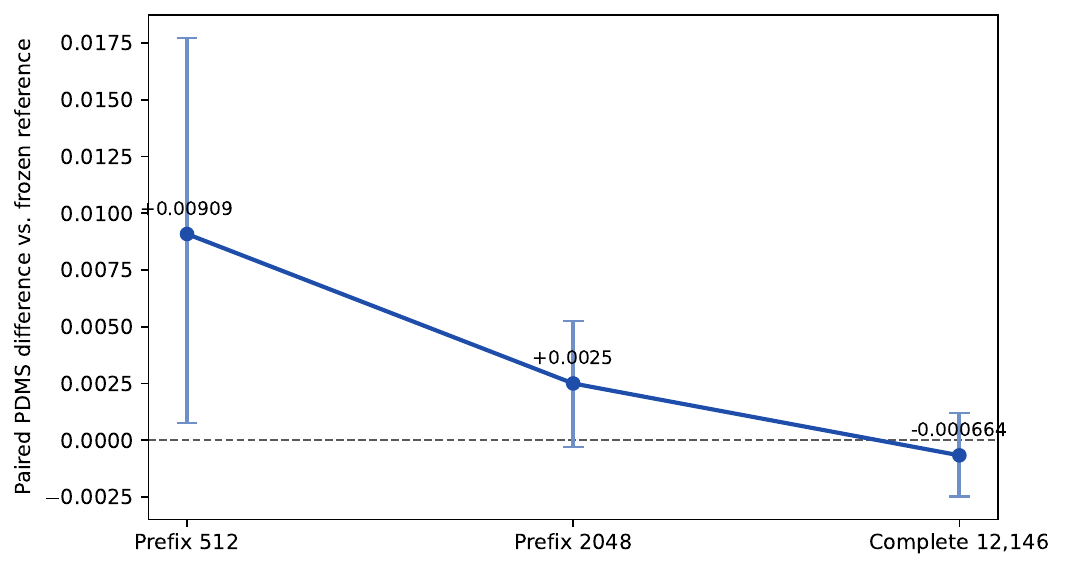}
\caption{Scale-expanded evaluation of the future-supervised candidate against the frozen reference. Error bars are 95\% percentile intervals from 10,000 paired scenario-bootstrap resamples. Only the prefix512 interval excludes zero; the complete-support mean is slightly negative.}
\label{fig:scale-curve}
\end{figure}

Figure~\ref{fig:scale-curve} visualizes the same scale expansion with paired uncertainty. The prefix512 interval excludes zero, whereas the larger supports do not. The line is descriptive because the three supports are nested evaluations and are not independent samples from a fitted trend.

\begin{figure}[H]
\begin{adjustwidth}{-\extralength}{0cm}
\centering
\subfloat[\centering\label{fig:paired-signs}]{\includegraphics[width=0.53\fulllength]{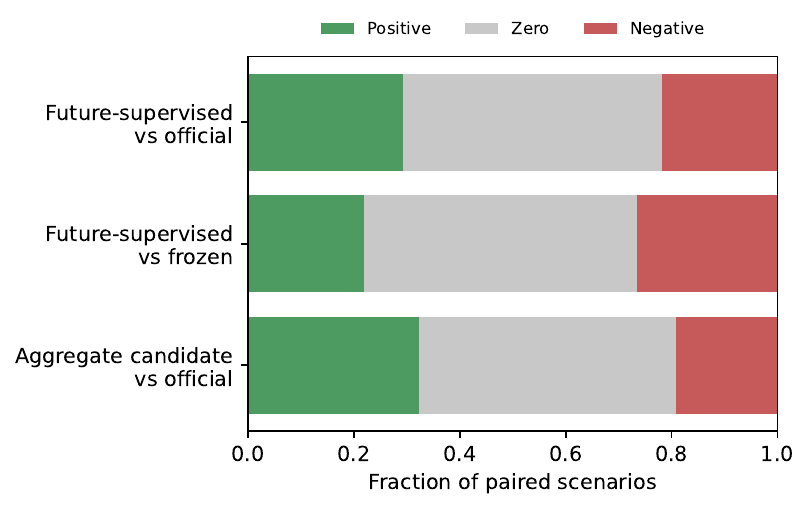}}
\hfill
\subfloat[\centering\label{fig:paired-ci}]{\includegraphics[width=0.43\fulllength]{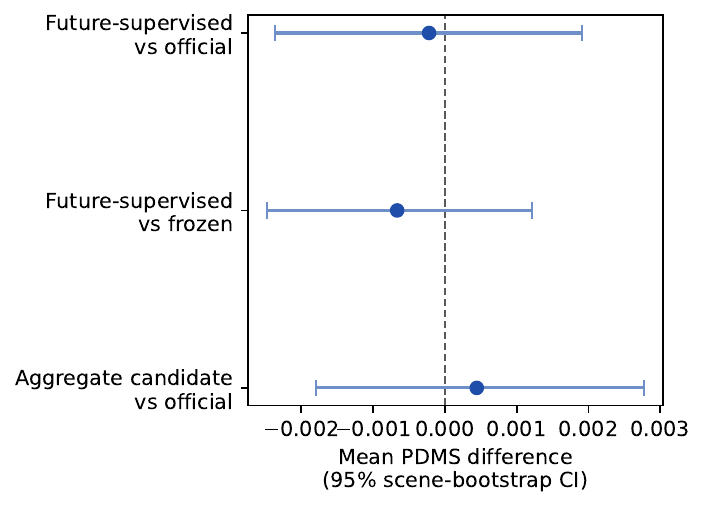}}
\end{adjustwidth}
\caption{Complete-support paired evidence. (\textbf{a}) Fractions of positive, tied, and negative scenario differences. (\textbf{b}) Mean paired PDMS differences with 95\% scene-bootstrap intervals. The intervals summarize the recorded scenario support and do not account for within-log dependence.}
\label{fig:paired-full}
\end{figure}

Figure~\ref{fig:paired-full} adds the paired scenario composition on complete support. Figure~\ref{fig:paired-signs} shows that every comparison contains a large mass of exact ties, whereas Figure~\ref{fig:paired-ci} shows that every mean interval crosses zero. This explains why a machine-precision aggregate sign and the fraction of locally improved scenarios are not interchangeable estimands.

\subsubsection{Test 4: Aggregate Utility and Critical-Component Regressions}

The exact paired full comparison in Table~\ref{tab:safety} illustrates why PDT retains both a meaningful-effect threshold and a component vector. The candidate and official checkpoint both have PDMS 0.937 at three significant digits. Their machine-precision difference is nominally positive, \(+4.44\times10^{-4}\), but its scene-bootstrap interval [\(-0.00179\), 0.00277] crosses zero, it is below 0.001, and no practical-effect threshold was preregistered for this comparison. It therefore does not support a meaningful aggregate-improvement claim. Ego progress and drivable-area compliance improve by 0.00666 and 0.00239, respectively, whereas collision, TTC, driving direction, and comfort decline. TTC has the largest mean regression at \(-0.00782\); the candidate therefore fails the zero-tolerance safety non-compensation rule used in this study.

This is not a claim that the candidate is unsafe in the real world; NAVSIM-v1 is a finite, non-reactive benchmark. It shows that a positive machine-precision sign, a practically meaningful aggregate effect, and no regression in critical components are three different claims. The first is observed, the second is indeterminate because no minimum meaningful effect was preregistered, and the third fails in this comparison.

\begin{table}[H]
\caption{Aggregate and component-level results for the exact paired full comparison. Values are means over 12,146 paired scenarios and are reported to approximately three significant digits.\label{tab:safety}}
\begin{adjustwidth}{-\extralength}{0cm}
\begin{tabularx}{\fulllength}{p{3.7cm}p{2.8cm}p{2.8cm}X}
\toprule
\textbf{Quantity} & \textbf{Candidate} & \textbf{Official} & \textbf{Candidate-minus-official verdict}\\
\midrule
PDMS & 0.937 & 0.937 & \(+4.44\times10^{-4}\); 95\% interval [\(-0.00179\), 0.00277]\\
Ego progress & 0.921 & 0.914 & \(+0.00666\)\\
Drivable-area compliance & 0.985 & 0.982 & \(+0.00239\)\\
No-at-fault collision & 0.989 & 0.991 & \(-0.00218\)\\
TTC within bound & 0.955 & 0.963 & \(-0.00782\)\\
Driving direction & 0.944 & 0.947 & \(-0.00276\)\\
Comfort & 0.999 & 0.999 & \(-1.65\times10^{-4}\)\\
\midrule
Reliability verdict & \multicolumn{3}{l}{Positive sign only; practical-effect indeterminate; safety non-compensation fail}\\
\bottomrule
\end{tabularx}
\end{adjustwidth}
\end{table}

\begin{figure}[H]
\centering
\includegraphics[width=0.94\textwidth]{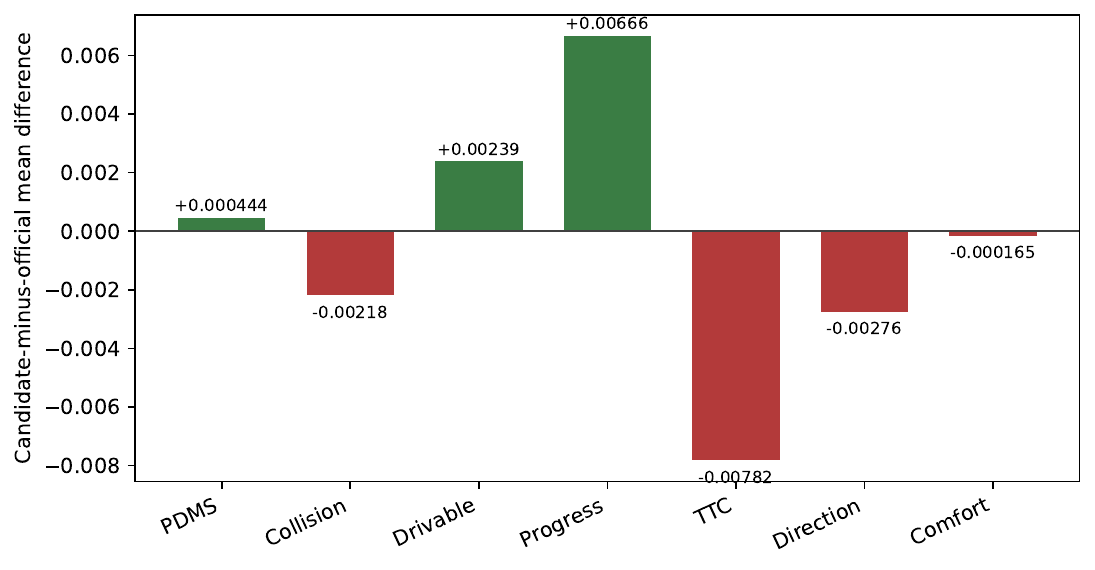}
\caption{Exact aggregate and component mean differences for the candidate versus the official checkpoint on 12,146 paired scenarios. Green denotes a positive mean difference and red a regression. The nominal aggregate sign does not satisfy the zero-tolerance non-compensation rule.}
\label{fig:component-directions}
\end{figure}

Figure~\ref{fig:component-directions} plots the exact mean differences recovered from the official paired CSVs. It makes the compensation structure visible: progress and drivable-area compliance increase, but their aggregate contribution coexists with larger-magnitude TTC and direction regressions.

\begin{figure}[H]
\centering
\includegraphics[width=0.80\textwidth]{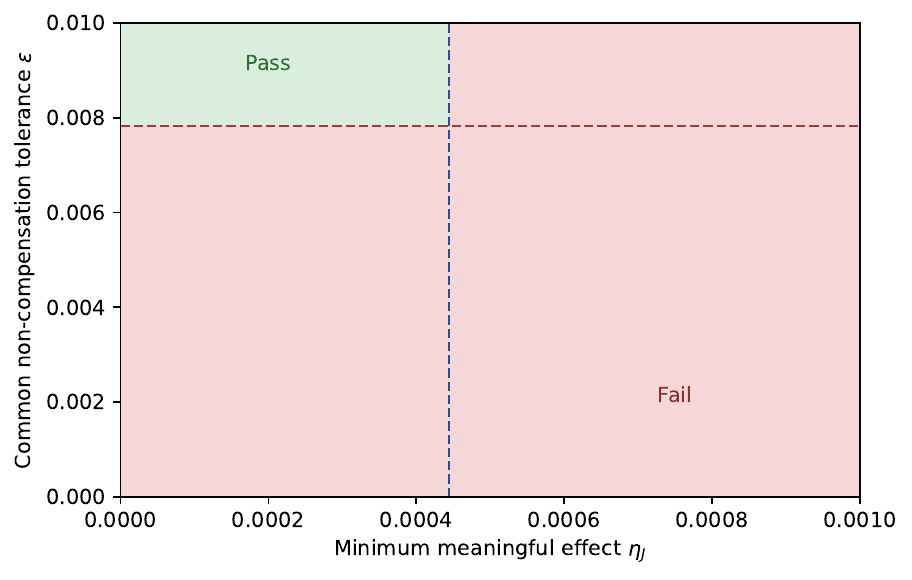}
\caption{Sensitivity of the aggregate reliability verdict to the minimum meaningful effect \(\eta_J\) and a common non-compensation tolerance \(\epsilon\). The dashed lines mark the observed aggregate difference and the minimum tolerance required to absorb the largest critical-component regression.}
\label{fig:threshold-sensitivity}
\end{figure}

The conclusion does not depend only on choosing a zero tolerance. Figure~\ref{fig:threshold-sensitivity} evaluates Equation~\eqref{eq:safety} over a grid of minimum meaningful effects \(\eta_J\) and a common component tolerance \(\epsilon\). A positive verdict is possible only when \(\eta_J\leq4.44\times10^{-4}\) and \(\epsilon\geq0.00782\). Thus any threshold requiring a larger aggregate effect, or any component tolerance stricter than the observed TTC regression, rejects the claim. The sensitivity surface displays the threshold choice explicitly instead of embedding it in one binary verdict.

\subsubsection{Test 5: Robustness of Broad Score Calibration}

Four calibration families---global future consensus, BCE, listwise, and set-wise conditional calibration---were evaluated before the final frozen-tensor screen. The exact screen trained 216 arms and evaluated four steps per arm, producing 864 seed-level outcomes and 432 two-seed configurations. Six individual seed-level outcomes had a positive sign; after pairing seeds, only four configurations had a positive average. None produced a strictly positive gain on both fixed validation halves across both seeds, and none reached the declared mean gain of 0.002.

Reporting the four nominal two-seed configurations alone would create a best-of-many positive result. The best set-wise configuration averages \(1.12\times10^{-4}\) across seeds, but all four seed-by-half gains are [\(4.49\times10^{-4}\), 0, 0, 0]. The repeatable linear configuration averages \(1.06\times10^{-4}\) and yields [\(2.13\times10^{-4}\), 0, \(2.13\times10^{-4}\), 0]. Under the declared family-level gate, the examined calibration family therefore does not demonstrate robust decision transfer. The candidate-oracle regret of 0.0125 shows that better choices exist inside the 32-proposal set, while the representative linear calibration reduces this regret by only \(1.06\times10^{-4}\). This identifies unresolved selection headroom; it does not prove a support bottleneck because no external comparison space was retained.

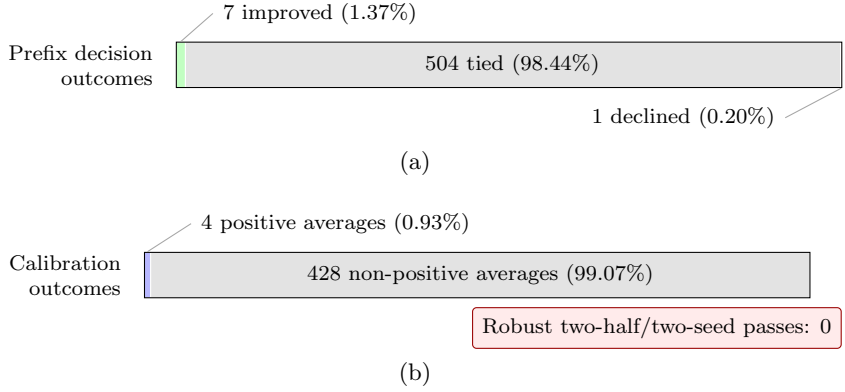
\begin{figure}[H]
\begin{adjustwidth}{-\extralength}{0cm}
\centering
\subfloat[\centering\label{fig:coverage-prefix}]{%
\begin{tikzpicture}[x=0.88cm,y=1cm,font=\small]
\node[anchor=east,align=right,font=\scriptsize] at (0,0) {Prefix decision\\outcomes};
\draw[fill=green!24,draw=white] (0.2,-0.30) rectangle (0.337,0.30);
\draw[fill=gray!22,draw=white] (0.337,-0.30) rectangle (10.181,0.30);
\draw[fill=red!22,draw=white] (10.181,-0.30) rectangle (10.2,0.30);
\draw[black] (0.2,-0.30) rectangle (10.2,0.30);
\node[font=\scriptsize] at (5.25,0) {504 tied (98.44\%)};
\draw[gray!70] (0.27,0.31) -- (0.75,0.68);
\node[anchor=west,font=\scriptsize] at (0.75,0.68) {7 improved (1.37\%)};
\draw[gray!70] (10.19,-0.31) -- (9.35,-0.70);
\node[anchor=east,font=\scriptsize] at (9.35,-0.70) {1 declined (0.20\%)};
\end{tikzpicture}}\\[3mm]
\subfloat[\centering\label{fig:coverage-calibration}]{%
\begin{tikzpicture}[x=0.88cm,y=1cm,font=\small]
\node[anchor=east,align=right,font=\scriptsize] at (0,0) {Calibration\\outcomes};
\draw[fill=blue!28,draw=white] (0.2,-0.30) rectangle (0.293,0.30);
\draw[fill=gray!22,draw=white] (0.293,-0.30) rectangle (10.2,0.30);
\draw[black] (0.2,-0.30) rectangle (10.2,0.30);
\node[font=\scriptsize] at (5.25,0) {428 non-positive averages (99.07\%)};
\draw[gray!70] (0.246,0.31) -- (0.9,0.68);
\node[anchor=west,font=\scriptsize] at (0.9,0.68) {4 positive averages (0.93\%)};
\node[draw=red!60!black,fill=red!8,rounded corners=1.5pt,font=\scriptsize] at (7.9,-0.70) {Robust two-half/two-seed passes: 0};
\end{tikzpicture}}
\end{adjustwidth}
\caption{Decision coverage and calibration robustness. (\textbf{a}) Exact outcome counts for the direct-utility candidate on the 512-record prefix. (\textbf{b}) Outcomes of the 432 two-seed configurations in the frozen-tensor screen; none of the four positive average gains satisfies the predefined robustness gate.}
\label{fig:coverage-robustness}
\end{figure}

Figure~\ref{fig:coverage-robustness} makes two forms of attrition explicit. The direct-utility candidate changes utility on only eight of 512 prefix records, and the calibration screen reduces 432 seed-paired configurations to four nominal average improvements and zero robust passes.

\subsection{Evaluation-Rule Ablation and Conclusion Reversals}
\label{subsec:ablation}

The main ablation removes one PDT claim rule at a time while holding the underlying artifacts fixed. This is an evaluation-rule ablation, not a new model-training experiment. Table~\ref{tab:ablation} shows that common shortcuts produce a positive conclusion that is rejected or bounded when the full transfer chain is restored.

\begin{table}[H]
\caption{Conclusion-reversal ablation. ``Numerically positive'' denotes a sign-only verdict and does not imply a practically meaningful effect.\label{tab:ablation}}
\begin{adjustwidth}{-\extralength}{0cm}
\begin{tabularx}{\fulllength}{p{3.8cm}p{4.0cm}p{3.2cm}X}
\toprule
\textbf{Removed PDT rule} & \textbf{Evidence viewed in isolation} & \textbf{Reduced-rule verdict} & \textbf{PDT verdict}\\
\midrule
Proxy-to-decision linkage & Component BCE \(0.705\rightarrow0.530\) & Positive proxy verdict & Fail: selected held PDM declines\\
Full-scale confirmation & Prefix512 \(+0.00909\); prefix2048 \(+0.00250\) & Positive subset verdict & Fail: complete-run PDMS is below the frozen reference and official checkpoint\\
Meaningful effect and safety non-compensation & Full PDMS nominally \(+4.44\times10^{-4}\) & Numerically positive & Practical-effect indeterminate; four critical/quality components decline\\
Family-level robustness & Best four positive averages among 432 two-seed configurations & Positive if cherry-picked & Fail: none passes both halves across both seeds\\
Artifact-completeness rule & Oracle headroom and unchanged selections in the boundary audit & Single-cause diagnosis & Indeterminate cause: proposal residual/ranking arrays are missing\\
\bottomrule
\end{tabularx}
\end{adjustwidth}
\end{table}

The ablation tests PDT's applied role: it changes whether evidence is used to expand a run, claim a safety-consistent improvement, or continue score-only calibration. The experiment does not show that every rule is mathematically necessary for every planner. Each rule prevents a documented conclusion error in the present case study.

\subsection{Visualization of Transfer Attrition}
\label{subsec:visualization}

\begin{figure}[H]
\centering
\begin{tikzpicture}[
  x=1cm,y=1cm,
  leftbox/.style={draw=green!45!black, fill=green!10, rounded corners=2pt,
    text width=4.10cm, minimum height=0.82cm, align=center, font=\scriptsize, inner sep=4pt},
  rightbox/.style={draw=red!55!black, fill=red!8, rounded corners=2pt,
    text width=4.10cm, minimum height=0.82cm, align=center, font=\scriptsize, inner sep=4pt},
  arr/.style={-{Latex[length=2mm]}, thick, gray!80}
]
\node[leftbox] (p1) at (0,3.0) {Better component proxy\\BCE decreases};
\node[rightbox] (n1) at (6.3,3.0) {Selected utility declines\\proxy transfer fails};
\draw[arr] (p1.east) -- (n1.west);
\node[leftbox] (p2) at (0,1.75) {Positive locked prefixes\\512 and 2048};
\node[rightbox] (n2) at (6.3,1.75) {Complete-run reversal\\deployment transfer fails};
\draw[arr] (p2.east) -- (n2.west);
\node[leftbox] (p3) at (0,0.50) {Nominal positive sign\\below 0.001};
\node[rightbox] (n3) at (6.3,0.50) {Meaningful effect unresolved;\\non-compensation fails};
\draw[arr] (p3.east) -- (n3.west);
\node[leftbox] (p4) at (0,-0.75) {Four positive averages\\among 432 configurations};
\node[rightbox] (n4) at (6.3,-0.75) {No two-half/two-seed pass\\robustness fails};
\draw[arr] (p4.east) -- (n4.west);
\draw[dashed,rounded corners=3pt,gray!80] (-2.45,-1.35) rectangle (2.45,3.70);
\draw[dashed,rounded corners=3pt,gray!80] (3.85,-1.35) rectangle (8.75,3.70);
\node[font=\small,fill=white,inner sep=2pt] at (0,3.70) {Local or reduced-rule evidence};
\node[font=\small,fill=white,inner sep=2pt] at (6.3,3.70) {PDT decision-level verdict};
\end{tikzpicture}
\caption{Observed conclusion reversals across the PDT chain. Positive local evidence is assigned to the transfer stage it supports and prevented from licensing a stronger downstream claim.}
\label{fig:reversals}
\end{figure}
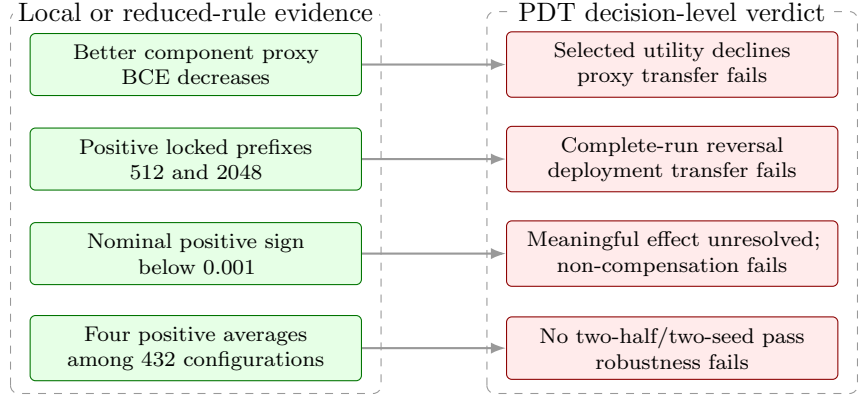

Figure~\ref{fig:reversals} summarizes the observed attrition between locally positive evidence and a reliable planning improvement claim. It shows verdict transitions; unavailable proposal-level arrays are not plotted. Three paths begin with a positive local observation and are rejected at selected utility, full-scale utility, or the critical component vector. The calibration path begins with isolated nominal improvements but fails replication across fixed halves and seeds.

\section{Discussion}
\label{sec:discussion}

\subsection{From Metric Reporting to an Evaluation Method}
\label{subsec:discussion_theory}

PDT provides an intermediate evaluation framework for future-aware planning signals. It is more specific than the principle of evaluating decisions rather than prediction error, yet it applies beyond a benchmark-specific checklist. Upstream proxy evidence supports a driving claim only when scoring, selection, support, and deployment conditions are observed or explicitly bounded. Proposition 1 defines proposal-wise activation surplus instead of using a global score-change norm. Proposition 2 separates the frequency and utility of decision switches, and Proposition 3 bounds score-only intervention under fixed candidate support. The reliability module limits the scope of conclusions drawn from these quantities.

The experiments evaluate the framework without requiring a positive model result. Several superficially positive premises change interpretation when the relevant transfer condition is tested. A lower proxy loss fails at selected utility; a prefix512 gain has a positive scene-bootstrap interval, but the prefix2048 and complete-support intervals cross zero; a nominal aggregate sign has interval [\(-0.00179\), 0.00277] and requires a component tolerance of at least 0.00782 to pass; and isolated calibration averages fail the robustness rule. The proposal-level replay verifies \(\Delta J=\rho\mu\) to floating-point precision and shows that most switches can be utility-neutral. Different outcomes can therefore be attributed to specific stages instead of being summarized as a single benchmark failure.

PDT separates three levels of evidence that are often compressed into a single claim. A \emph{learning claim} tests whether the future target can be predicted. A \emph{planning improvement claim} tests whether the induced scores change selected trajectories with positive utility. A \emph{deployment claim} tests whether that benefit survives full support and critical-component constraints. Passing an earlier level does not entail the next. The terms ``improved representation,'' ``improved selection,'' and ``reliable planning gain'' therefore refer to different evidential strengths.

\subsection{Implications for Future-Aware Planner Development}
\label{subsec:discussion_practice}

The first unresolved transfer stage determines the next experiment. If a held future objective does not improve, further benchmark evaluation is premature and effort should return to the representation target. If the representation improves but proposal ordering does not, proposal-wise utilities, score correlations, and margin-normalized perturbations should be inspected before adding more training epochs. If scores change but selections do not, the relevant quantities are the top-1/top-2 margin distribution and switch coverage; another proxy-loss measurement adds little information. If selection regret is small relative to support regret, further score calibration has limited leverage and the candidate generator should be expanded or diversified. If a local decision gain fails scale, safety, or sequential comparability gates, the claim should be bounded and the failed deployment condition investigated without reopening unrestricted checkpoint search.

The case study illustrates each of these choices. The early decision-boundary audit contains oracle headroom but no observed switch, so its missing proposal arrays prevent a single-cause diagnosis. The later frozen-tensor replay resolves that measurement gap on a separate support: the repeatable linear calibration changes four of 64 decisions, with \(\rho=0.0625\) and \(\mu=0.00170\), but three of the four switches are utility-neutral. The direct-utility candidate affects only eight of 512 prefix records. The future-supervised candidate warrants scale expansion because its prefix512 interval is positive, but uncertainty includes zero at prefix2048 and complete support. Finally, the aggregate-score and calibration experiments show that a machine-precision sign, a practically meaningful effect, component non-compensation, and family-level robustness are separate objectives.

For implementation, PDT requires a minimal logging interface independent of planner architecture: stable candidate identifiers, proposal-wise scores, selected indices, top-1/top-2 margins, per-candidate utility where feasible, aggregate and component outcomes, evaluator-state metadata, and immutable support keys. Persisting these fields during training and evaluation makes the transfer chain auditable without redesigning the planner. It also reduces the risk that a later analysis becomes indeterminate because a checkpoint, residual array, or evaluator state was not retained.

\subsection{Limitations and Future Work}
\label{subsec:limitations}

The current evidence has four main limitations. First, PDT is evaluated on one proposal-based future-aware planning family and NAVSIM-v1. The case study demonstrates diagnostic usefulness, but it does not show cross-architecture or cross-benchmark invariance. Second, the framework was synthesized retrospectively from a sequence of experiments. Several campaign-level thresholds and fixed splits were locked before their corresponding outcomes, but the complete taxonomy was not preregistered. Third, proposal arrays were recovered for the 64-token frozen calibration support but not for the early four-token audit or the complete 12,146-record runs. The exported complete-run CSVs also lack log identifiers. We can therefore report proposal-level margin and switch diagnostics on the frozen validation support and scenario-level paired intervals on complete support, but not log-clustered intervals, complete-run switch-conditioned utility, or an external support-regret decomposition. Fourth, non-reactive benchmark evidence cannot demonstrate real-world safety or closed-loop causal effects under interactive traffic.

Future work should therefore evaluate PDT prospectively across sparse, diffusion-based, and world-model-based planners with a common logging interface. A preregistered study should freeze the proxy, utility, safety tolerances, statistical unit, candidate support, and scale-expansion gates before outcomes are opened. Route- or log-level resampling can then quantify clustered uncertainty in \(\rho\), \(\mu\), and paired utility, while reactive or closed-loop evaluation can test whether the same failure locations remain informative when the policy changes subsequent observations. Controlled support-expansion experiments are especially important: adding diverse, feasible candidates while holding the scorer fixed would identify whether the residual regret is caused by selection or candidate support.

\section{Conclusions}
\label{sec:conclusion}

This paper introduced the Proxy-to-Decision Transfer Framework for evaluating future-aware proposal-based autonomous-driving planners. PDT analyzes decision dependencies across multiple stages and combines two modules: Decision-Transfer Decomposition localizes failures across representation, scoring, selection, and candidate support, whereas Reliability-Constrained Validation limits positive claims through exact pairing, scale expansion, critical-component non-compensation, sequential comparability, and family-level robustness. The framework analyzes how proxy evidence affects planning decisions without introducing a new planner architecture or scalar benchmark.

The NAVSIM-v1 case study demonstrates why these distinctions matter. Proxy improvement coexists with lower selected utility; a proposal replay closes \(\Delta J=\rho\mu\) exactly but reveals low decision coverage and mostly utility-neutral switches; a positive prefix512 interval becomes unresolved on larger supports; two aggregate PDMS values are identical at three significant digits while four critical or quality components decline; and none of 432 two-seed configurations passes the predefined robustness rule. These findings do not indicate that future-aware learning is ineffective; they show that its benefits should be evaluated at the decision level and under explicit support, uncertainty, and reliability conditions. Reporting pass, fail, and indeterminate cases at specific transfer stages provides a reproducible basis for model selection, experiment design, and appropriately scoped conclusions.

\section*{Data and Code Availability}
NAVSIM is publicly available from its cited source. The arXiv ancillary files contain the minimal retained evidence supporting the reported statistics: paired scenario-level metrics, frozen proposal tensors, configuration locks, proposal-diagnostic outputs, derived tables, deterministic analysis scripts, and SHA-256 hashes. Large public benchmark assets and upstream model weights are excluded and should be obtained from their original repositories.


\begin{thebibliography}{99}
\bibitem{hu2023uniad} Hu, Y.; Yang, J.; Chen, L.; et al. Planning-Oriented Autonomous Driving. In \emph{Proceedings of the IEEE/CVF Conference on Computer Vision and Pattern Recognition}; 2023; pp. 17853--17862.
\bibitem{jiang2023vad} Jiang, B.; Chen, S.; Xu, Q.; et al. VAD: Vectorized Scene Representation for Efficient Autonomous Driving. In \emph{Proceedings of the IEEE/CVF International Conference on Computer Vision}; 2023; pp. 8340--8350.
\bibitem{wang2024drivewm} Wang, Y.; He, J.; Fan, L.; Li, H.; Chen, Y.; Zhang, Z. Driving into the Future: Multiview Visual Forecasting and Planning with World Model for Autonomous Driving. In \emph{Proceedings of the IEEE/CVF Conference on Computer Vision and Pattern Recognition}; 2024; pp. 14749--14759.
\bibitem{li2025law} Li, Z.; et al. Enhancing End-to-End Autonomous Driving with Latent World Model. In \emph{Proceedings of the International Conference on Learning Representations}; 2025. Available online: \url{https://openreview.net/forum?id=fd2u60ryG0} (accessed on 2 September 2026).
\bibitem{zheng2025world4drive} Zheng, Y.; Yang, P.; Xing, Z.; et al. World4Drive: End-to-End Autonomous Driving via Intention-Aware Physical Latent World Model. In \emph{Proceedings of the IEEE/CVF International Conference on Computer Vision}; 2025; pp. 28632--28642.
\bibitem{li2025wote} Li, Y.; et al. End-to-End Driving with Online Trajectory Evaluation via BEV World Model. In \emph{Proceedings of the IEEE/CVF International Conference on Computer Vision}; 2025.
\bibitem{wang2026drivejepa} Wang, L.; Yang, Z.; Bai, C.; et al. Drive-JEPA: Video JEPA Meets Multimodal Trajectory Distillation for End-to-End Driving. \emph{arXiv} \textbf{2026}, arXiv:2601.22032.
\bibitem{dauner2024navsim} Dauner, D.; Hallgarten, M.; Li, T.; et al. NAVSIM: Data-Driven Non-Reactive Autonomous Vehicle Simulation and Benchmarking. \emph{Advances in Neural Information Processing Systems} \textbf{2024}, \emph{37}. \href{https://doi.org/10.52202/079017-0902}{https://doi.org/10.52202/079017-0902}.
\bibitem{li2024egostatus} Li, Z.; Yu, Z.; Lan, S.; et al. Is Ego Status All You Need for Open-Loop End-to-End Autonomous Driving? In \emph{Proceedings of the IEEE/CVF Conference on Computer Vision and Pattern Recognition}; 2024; pp. 14864--14873.
\bibitem{dauner2023misconceptions} Dauner, D.; Hallgarten, M.; Geiger, A.; Chitta, K. Parting with Misconceptions about Learning-Based Vehicle Motion Planning. In \emph{Proceedings of the Conference on Robot Learning}; PMLR, 2023; Volume 229.
\bibitem{jia2024bench2drive} Jia, X.; et al. Bench2Drive: Towards Multi-Ability Benchmarking of Closed-Loop End-to-End Autonomous Driving. \emph{Advances in Neural Information Processing Systems} \textbf{2024}, \emph{37}.
\bibitem{elmachtoub2022spo} Elmachtoub, A.N.; Grigas, P. Smart “Predict, then Optimize”. \emph{Management Science} \textbf{2022}, \emph{68}, 9--26. \href{https://doi.org/10.1287/mnsc.2020.3922}{https://doi.org/10.1287/mnsc.2020.3922}.
\bibitem{wilder2019decision} Wilder, B.; Dilkina, B.; Tambe, M. Melding the Data-Decisions Pipeline: Decision-Focused Learning for Combinatorial Optimization. \emph{Proceedings of the AAAI Conference on Artificial Intelligence} \textbf{2019}, \emph{33}, 1658--1665. \href{https://doi.org/10.1609/aaai.v33i01.33011658}{https://doi.org/10.1609/aaai.v33i01.33011658}.
\bibitem{mandi2022ranking} Mandi, J.; Bucarey, V.; Tchomba, M.M.; Guns, T. Decision-Focused Learning: Through the Lens of Learning to Rank. In \emph{Proceedings of the 39th International Conference on Machine Learning}; PMLR, 2022; Volume 162.
\bibitem{dwork2015adaptive} Dwork, C.; Feldman, V.; Hardt, M.; Pitassi, T.; Reingold, O.; Roth, A. Generalization in Adaptive Data Analysis and Holdout Reuse. \emph{Advances in Neural Information Processing Systems} \textbf{2015}, \emph{28}.
\bibitem{codevilla2018cil} Codevilla, F.; M\"uller, M.; L\'opez, A.M.; Koltun, V.; Dosovitskiy, A. End-to-End Driving via Conditional Imitation Learning. In \emph{Proceedings of the IEEE International Conference on Robotics and Automation}; 2018; pp. 4693--4700. \href{https://doi.org/10.1109/ICRA.2018.8460487}{https://doi.org/10.1109/ICRA.2018.8460487}.
\bibitem{zeng2019nmp} Zeng, W.; Luo, W.; Suo, S.; Sadat, A.; Yang, B.; Casas, S.; Urtasun, R. End-to-End Interpretable Neural Motion Planner. In \emph{Proceedings of the IEEE/CVF Conference on Computer Vision and Pattern Recognition}; 2019; pp. 8652--8661. \href{https://doi.org/10.1109/CVPR.2019.00886}{https://doi.org/10.1109/CVPR.2019.00886}.
\bibitem{casas2021mp3} Casas, S.; Sadat, A.; Urtasun, R. MP3: A Unified Model to Map, Perceive, Predict and Plan. In \emph{Proceedings of the IEEE/CVF Conference on Computer Vision and Pattern Recognition}; 2021; pp. 14398--14407. \href{https://doi.org/10.1109/CVPR46437.2021.01417}{https://doi.org/10.1109/CVPR46437.2021.01417}.
\bibitem{hu2022stp3} Hu, S.; Chen, L.; Wu, P.; Li, H.; Yan, J.; Tao, D. ST-P3: End-to-End Vision-Based Autonomous Driving via Spatial--Temporal Feature Learning. In \emph{Proceedings of the European Conference on Computer Vision}; 2022; pp. 533--549. \href{https://doi.org/10.1007/978-3-031-19839-7_31}{https://doi.org/10.1007/978-3-031-19839-7\_31}.
\bibitem{renz2022plant} Renz, K.; Chitta, K.; Mercea, O.-B.; Koepke, A.S.; Akata, Z.; Geiger, A. PlanT: Explainable Planning Transformers via Object-Level Representations. In \emph{Proceedings of the 6th Conference on Robot Learning}; PMLR, 2023; Volume 205, pp. 459--470.
\bibitem{chitta2023transfuser} Chitta, K.; Prakash, A.; Jaeger, B.; Yu, Z.; Renz, K.; Geiger, A. TransFuser: Imitation with Transformer-Based Sensor Fusion for Autonomous Driving. \emph{IEEE Trans. Pattern Anal. Mach. Intell.} \textbf{2023}, \emph{45}, 12878--12895. \href{https://doi.org/10.1109/TPAMI.2022.3200245}{https://doi.org/10.1109/TPAMI.2022.3200245}.
\bibitem{chen2024e2e} Chen, L.; Wu, P.; Chitta, K.; Jaeger, B.; Geiger, A.; Li, H. End-to-End Autonomous Driving: Challenges and Frontiers. \emph{IEEE Trans. Pattern Anal. Mach. Intell.} \textbf{2024}, \emph{46}, 10164--10183. \href{https://doi.org/10.1109/TPAMI.2024.3435937}{https://doi.org/10.1109/TPAMI.2024.3435937}.
\bibitem{sun2024sparsedrive} Sun, W.; Lin, X.; Shi, Y.; Zhang, C.; Wu, H.; Zheng, S. SparseDrive: End-to-End Autonomous Driving via Sparse Scene Representation. In \emph{Proceedings of the IEEE International Conference on Robotics and Automation}; 2025; pp. 8795--8801. \href{https://doi.org/10.1109/ICRA55743.2025.11128800}{https://doi.org/10.1109/ICRA55743.2025.11128800}.
\bibitem{liao2025diffusiondrive} Liao, B.; Chen, S.; Yin, H.; Jiang, B.; Wang, C.; Yan, S.; et al. DiffusionDrive: Truncated Diffusion Model for End-to-End Autonomous Driving. In \emph{Proceedings of the IEEE/CVF Conference on Computer Vision and Pattern Recognition}; 2025; pp. 12037--12047. \href{https://doi.org/10.1109/CVPR52734.2025.01124}{https://doi.org/10.1109/CVPR52734.2025.01124}.
\bibitem{song2025momad} Song, Z.; Jia, C.; Liu, L.; Pan, H.; Zhang, Y.; Wang, J.; et al. Don't Shake the Wheel: Momentum-Aware Planning in End-to-End Autonomous Driving. In \emph{Proceedings of the IEEE/CVF Conference on Computer Vision and Pattern Recognition}; 2025; pp. 22432--22441. \href{https://doi.org/10.1109/CVPR52734.2025.02089}{https://doi.org/10.1109/CVPR52734.2025.02089}.
\bibitem{wu2026mfpad} Wu, Y.; Hu, Q.; Lei, A.; Song, Z. MFPAD: Memory--Forgetting Planning for Long-Horizon End-to-End Autonomous Driving. \emph{IEEE Latin Am. Trans.} \textbf{2026}, \emph{24}, 753--764. \href{https://doi.org/10.1109/TLA.2026.11577664}{https://doi.org/10.1109/TLA.2026.11577664}.
\bibitem{hu2021freespace} Hu, P.; Huang, A.; Dolan, J.; Held, D.; Ramanan, D. Safe Local Motion Planning with Self-Supervised Freespace Forecasting. In \emph{Proceedings of the IEEE/CVF Conference on Computer Vision and Pattern Recognition}; 2021; pp. 12727--12736. \href{https://doi.org/10.1109/CVPR46437.2021.01254}{https://doi.org/10.1109/CVPR46437.2021.01254}.
\bibitem{min2024driveworld} Min, C.; Zhao, D.; Xiao, L.; Zhao, J.; Xu, X.; Zhu, Z.; et al. DriveWorld: 4D Pre-Trained Scene Understanding via World Models for Autonomous Driving. In \emph{Proceedings of the IEEE/CVF Conference on Computer Vision and Pattern Recognition}; 2024; pp. 15522--15533. \href{https://doi.org/10.1109/CVPR52733.2024.01470}{https://doi.org/10.1109/CVPR52733.2024.01470}.
\bibitem{zheng2024occworld} Zheng, W.; Chen, W.; Huang, Y.; Zhang, B.; Duan, Y.; Lu, J. OccWorld: Learning a 3D Occupancy World Model for Autonomous Driving. In \emph{Proceedings of the European Conference on Computer Vision}; 2024; pp. 55--72. \href{https://doi.org/10.1007/978-3-031-72624-8_4}{https://doi.org/10.1007/978-3-031-72624-8\_4}.
\bibitem{wang2023drivedreamer} Wang, X.; Zhu, Z.; Huang, G.; Chen, X.; Lu, J. DriveDreamer: Towards Real-World-Driven World Models for Autonomous Driving. \emph{arXiv} \textbf{2023}, arXiv:2309.09777.
\bibitem{hu2023gaia} Hu, A.; Russell, L.; Yeo, H.; Murez, Z.; Fedoseev, G.; Kendall, A.; Shotton, J.; Corrado, G. GAIA-1: A Generative World Model for Autonomous Driving. \emph{arXiv} \textbf{2023}, arXiv:2309.17080.
\bibitem{gao2024vista} Gao, S.; Yang, J.; Chen, L.; Chitta, K.; Qiu, Y.; Geiger, A.; Zhang, J.; Li, H. Vista: A Generalizable Driving World Model with High Fidelity and Versatile Controllability. \emph{Advances in Neural Information Processing Systems} \textbf{2024}, \emph{37}. \href{https://doi.org/10.52202/079017-2906}{https://doi.org/10.52202/079017-2906}.
\bibitem{fu2026prodrive} Fu, C.; Gan, S.; Ouyang, Z.; Rui, Y.; Chi, X.; Han, S.; Wang, J.; Zhang, H. ProDrive: Proactive Planning for Autonomous Driving via Ego--Environment Co-Evolution. \emph{arXiv} \textbf{2026}, arXiv:2604.25329.
\bibitem{caesar2020nuscenes} Caesar, H.; Bankiti, V.; Lang, A.H.; Vora, S.; Liong, V.E.; Xu, Q.; et al. nuScenes: A Multimodal Dataset for Autonomous Driving. In \emph{Proceedings of the IEEE/CVF Conference on Computer Vision and Pattern Recognition}; 2020; pp. 11618--11628. \href{https://doi.org/10.1109/CVPR42600.2020.01164}{https://doi.org/10.1109/CVPR42600.2020.01164}.
\bibitem{caesar2021nuplan} Caesar, H.; Kabzan, J.; Tan, K.S.; Fong, W.K.; Wolff, E.M.; Lang, A.H.; Fletcher, L.; Beijbom, O.; Omari, S. nuPlan: A Closed-Loop ML-Based Planning Benchmark for Autonomous Vehicles. \emph{arXiv} \textbf{2021}, arXiv:2106.11810.
\bibitem{yang2024drivearena} Yang, X.; Wen, L.; Wei, T.; Ma, Y.; Mei, J.; Li, X.; et al. DriveArena: A Closed-Loop Generative Simulation Platform for Autonomous Driving. In \emph{Proceedings of the IEEE/CVF International Conference on Computer Vision}; 2025; pp. 26933--26943.
\bibitem{cao2025pseudosim} Cao, W.; Hallgarten, M.; Li, T.; Dauner, D.; Gu, X.; Wang, C.; et al. Pseudo-Simulation for Autonomous Driving. \emph{arXiv} \textbf{2025}, arXiv:2506.04218.
\bibitem{zhai2023rethinking} Zhai, J.-T.; Feng, Z.; Du, J.; Mao, Y.; Liu, J.-J.; Tan, Z.; Zhang, Y.; Ye, X.; Wang, J. Rethinking the Open-Loop Evaluation of End-to-End Autonomous Driving in nuScenes. \emph{arXiv} \textbf{2023}, arXiv:2305.10430.
\end{thebibliography}
\end{document}